\documentclass[conference, 10pt]{IEEEtran}
\usepackage{multirow}% http://ctan.org/pkg/multirow
\usepackage[font=scriptsize,caption=false,labelsep=space]{subfig}
\usepackage{mathrsfs}
\usepackage{graphicx,float,wrapfig,epstopdf,amsmath}
\usepackage[]{algorithmicx}
\usepackage{algpseudocode,algorithm}
\usepackage{balance}

\usepackage{amsmath}
\usepackage{amssymb}
\usepackage{amsthm}
\usepackage{graphicx}
\usepackage{epstopdf}
\usepackage{times}
\usepackage{textcomp}
\usepackage{color}
\usepackage{url}
\usepackage{cite}
\usepackage{multirow}
\usepackage[table]{xcolor}% http://ctan.org/pkg/xcolor
\usepackage{threeparttable}
\graphicspath{{./figures/}}

\DeclareMathOperator*{\minimize}{minimize} % thin space, limits underneath in 
\DeclareMathOperator*{\subjectto}{subject \hspace{3pt} to:\hspace{3pt}} % thin space, limits

\newcommand{\subparagraph}{}
\usepackage[compact]{titlesec}
\titlespacing{\section}{0pt}{2ex}{1ex}
\titlespacing{\subsection}{0pt}{1ex}{0ex}
\titlespacing{\subsubsection}{0pt}{0.5ex}{0ex}

\begin{document}
	
	% paper title
	% can use linebreaks \\ within to get better formatting as desired
	\title{ Hybrid Federated and Centralized Learning}
	
	\author{
		\IEEEauthorblockN{Ahmet~M. Elbir$^{1}$, 
			Sinem Coleri$^{1}$ and Kumar Vijay Mishra$^2$
		}%
		\IEEEauthorblockA{$^{1}$ Department of Electrical and Electronics Engineering, Ko{c} University, Istanbul, Turkey\\ 
			$^2$ United States CCDC Army Research Laboratory, Adelphi, MD 20783 USA\\
			E-mail: {ahmetmelbir@gmail.com, scoleri@ku.edu.tr, kumarvijay-mishra@uiowa.edu}
			%			\vspace{-8pt}
		}
	}
	
	% make the title area
	\maketitle
	
	%\vspace{-10pt}
	\begin{abstract}
		Many of the machine learning (ML) tasks are focused on centralized learning (CL), which requires the transmission of local datasets from the clients to a parameter server (PS) leading to a huge communication overhead. Federated learning (FL) overcomes this issue by allowing the clients to send only the model updates to the PS instead of the whole dataset. In this way, FL brings the learning to edge level, wherein powerful computational resources are required on the client side. This requirement may not always be satisfied because of diverse computational capabilities of edge devices. We address this through a novel hybrid federated and centralized learning (HFCL) framework to effectively train a learning model by exploiting the computational capability of the clients. In HFCL, only the clients who have sufficient resources employ FL; the remaining clients resort to CL by transmitting their local dataset to PS. This allows all the clients to collaborate on the learning process regardless of their computational resources. We also propose a sequential data transmission approach with HFCL (HFCL-SDT) to reduce the training duration. The proposed HFCL frameworks outperform previously proposed non-hybrid FL (CL) based schemes in terms of learning accuracy (communication overhead) since all the clients collaborate on the learning process with their datasets regardless of their computational resources.
	\end{abstract}
	\begin{IEEEkeywords}
		Machine learning, federated learning, centralized learning, edge intelligence,  edge efficiency.
	\end{IEEEkeywords}

	\section{Introduction}
	Machine learning (ML) has emerged as a promising engineering for future technologies such as internet of things (IoT), autonomous driving and next-generation wireless communications \cite{deepLearningScience,hodge2021deep}. These applications require massive data processing and abstraction by a learning model, often an artificial neural network (ANN), by extracting the features from the raw data and providing a ``meaning'' to the input via constructing model-free data mapping with huge number of learnable parameters~\cite{elbir2020cognitive,elbir2020DL4IRS_survey}. The implementation of these learning models demands powerful computational resources, such as graphics processing units (GPUs). Therefore, huge learning models, massive amount of training data, and powerful computation infrastructure are the  main driving factors of the success of ML algorithms \cite{survey_DL_Scalable}.
	
	Implementations of ML usually focus on centralized learning (CL) algorithms, where a powerful ANN is trained at a parameter server (PS) \cite{deepLearningScience,edgeWireless, fl_spm_federatedLearning}. This inherently assumes the availability of data at the PS. In case of wireless edge devices (clients), transmitting the collected data to the PS in a reliable manner is expensive in terms of energy and bandwidth, thereby introducing delays and possible infringements of client privacy \cite{FL_Gunduz}. For example, in long-term evolution (LTE) networks, a single frame of $5$ MHz bandwidth and $10$ ms duration may carry only $6000$ complex symbols~\cite{FL_gunduz_fading}, whereas the size of the whole dataset could be hundreds of thousands symbols~\cite{elbirQuantizedCNN2019}. As a result, CL-based techniques require huge bandwidths and communications overhead during training. 
	
	As a practically viable alternative to CL-based training, federated learning (FL) has been proposed to exploit the processing capability of the edge devices and the local datasets of the clients~\cite{fl_spm_federatedLearning,fl_By_Google}. In FL, rather than transmitting local datasets to the PS, the clients send the model updates (gradients) to collaboratively train the learning model. The collected model updates are aggregated at the PS and then sent back to the clients to further update the learning parameters iteratively. Compared to CL, FL provides less communications overhead with a slight prediction loss because of insufficient gradient components and data corruptions during wireless gradient transmission. 
	
	Recently, FL has been applied to %image classification~\cite{fl_By_Google,FL_Gunduz,FL_overtheair}, speech recognition~\cite{fl_speechRecognition} and 
	wireless communications~\cite{elbir2020FL_HB,elbir2020_FL_CE}, including architectures such as cellular networks~\cite{irs_FL_BF_fromChannel,FL_Bennis3,FL_gunduz_fading}, vehicular networks~\cite{elbir2020federated}, unmanned aerial vehicles~\cite{FL_Bennis2} and IoT networks~\cite{fl_IoT}. Here, FL architectures relied on the fact that all of the clients were capable of gradient computation, often using powerful parallel processing units. However, in practice, considering the diversity of the devices with different computational capabilities - such as mobile phones, vehicular components and IoT devices - this requirement cannot be met. A possible solution lies in a hybrid learning technique that benefits from both CL and FL. Here, the devices incapable of sufficient computation power deploy CL while the rest use FL. To this end, the client selection algorithms proposed in \cite{FL_vehicular1,fl_clientSelection} do not involve a collaboration of all the devices; instead, only the trusted clients~\cite{FL_vehicular1} or the ones  with sufficient computational resources~\cite{fl_clientSelection} are selected for FL-based training.  
	
\begin{figure}[t]
		\centering
		{\includegraphics[draft=false,scale=.055]{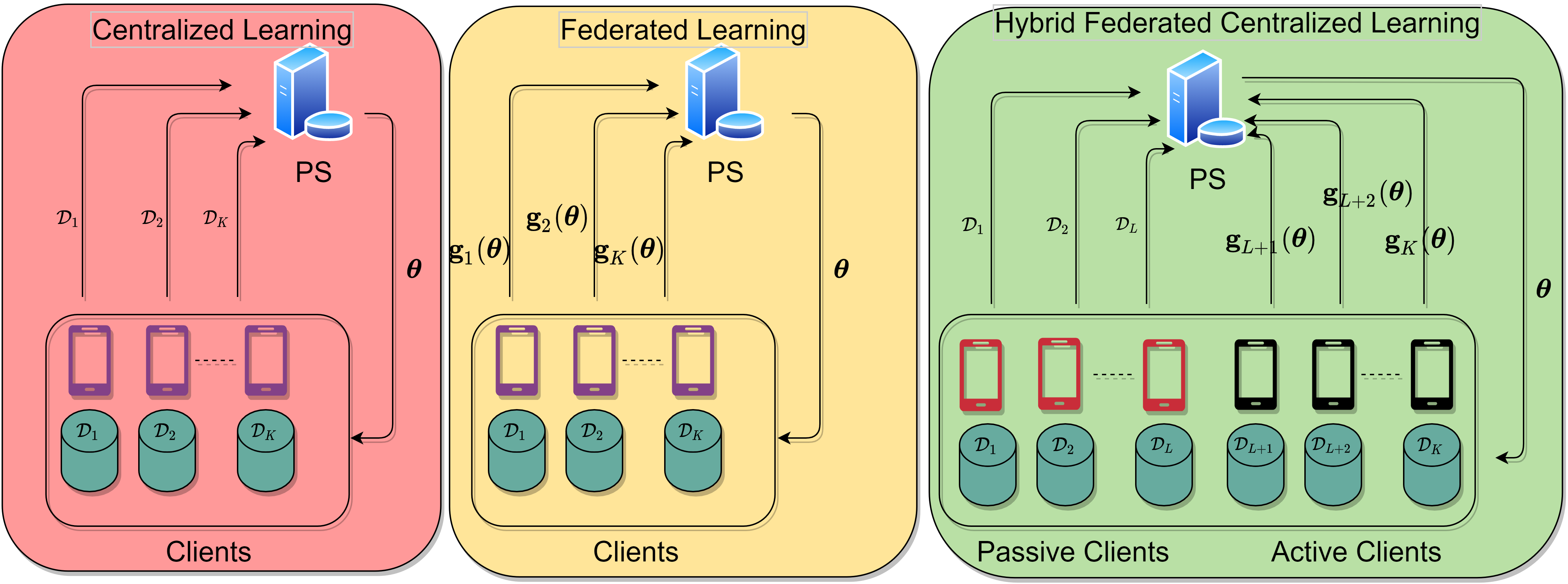}	\label{fig_Diaga} } 
		\caption{In a CL framework (left), all clients transmit their datasets to the PS. On the other hand, the datasets are preserved at the clients in FL (center) and only model parameters are sent to the PS. In HFCL (right), the clients are grouped into active and passive depending on their computational capability to either perform CL and FL, respectively.
		}
		%			\vspace*{-5mm}
		\label{fig_Diag}
	\end{figure}

	In this paper, we introduce a hybrid FL and CL (HFCL) framework to effectively train a learning model where the edge devices collaborate on the learning process by exploiting their computational capabilities (Fig.~\ref{fig_Diag}). %To the best of our knowledge this is the first work to employ a hybrid architecture exploiting the hardware capability of the edge devices. 
	In the beginning, we designate the clients as \emph{passive} (CL) or \emph{active} (FL) depending on their computational power. Then, the active clients transmit the gradient information to the PS based on their local dataset. On the other hand, the passive clients transmit their dataset to the PS, which computes the corresponding gradients data on behalf of them. At the PS, the computed gradients are utilized to aggregate the model parameters, which are then sent back to active devices. The challenge for HFCL is the wait required by the active clients when the passive clients complete their data set transmission at the beginning of the training. To mitigate this problem, we propose a sequential dataset transmission (SDT) approach  where the passive clients do not send the entire local dataset at once. Rather the local dataset is divided into smaller blocks so that both active/passive devices perform gradient/data transmission during the same communication exchange. We evaluate the performance of the proposed approaches on MNIST dataset~\cite{mnistlecun2010mnist} and show that both techniques provide higher learning performance than FL while introducing a slight increase in the communication overhead arising from dataset transmission. However, this overhead is still less than that of CL.
	
	%	The rest of the paper is organized as follows. In Section II, the preliminaries on CL and FL are given. We introduce the proposed HFCL and HFCL-SDT approaches in Section III and Section IV, respectively. The numerical simulations are presented in Section V and the paper is finalized with conclusions in Section VI.
	
	\textit{Notation:} Throughout the paper,  we denote the vectors and matrices by boldface lower and upper case symbols, respectively. In case of a vector $\mathbf{a}$, $[\mathbf{a}]_{i}$ represents its $i$-th element. For a matrix $\mathbf{A}$, $[\mathbf{A}]_{i,j}$ denotes the $(i,j)$-th entry. The $\mathbf{I}_N$ is the identity matrix of size $N\times N$;  $\|\cdot\|_\mathcal{F}$ is the Frobenius norm; The notation expressing a convolutional layer with $N$ filters/channels of size $D\times D$ is given by  $N$@$ D\times D$. 
	%\textcolor{red}{Notations para: Frobenius norm, etc.}
	
	\section{Desiderata on CL and FL}
	
	%%-----------------------------------------------------
	%%-----------------------------------------------------
	
	%	\section{Problem Description}
	Consider a scenario, wherein $K$ edge devices collaborate on solving an optimization problem through ML. Define $\mathcal{D}_k$ to be the local dataset of the $k$-th client so that the whole dataset is $\mathcal{D} = \bigcup_{k\in \mathcal{K}} \mathcal{D}_k$, where $\mathcal{K} = \{1,\dots,K\}$. Denote the input and output tuple $\mathcal{D}_i$ of the $i$-th element of the dataset $\mathcal{D}$ by $\mathcal{X}_i$ and $\mathcal{Y}_i$, respectively, such that $\mathcal{D}_i = (\mathcal{X}_i,\mathcal{Y}_i)$ and $\textsf{D}= |\mathcal{D}|$ is the number of input-output pairs. The CL-based training considers a learning model parameterized by $\boldsymbol{\theta}\in \mathbb{R}^P$ and optimized via
	\begin{align}
	\label{lossML}
	\minimize_{\boldsymbol{\theta}} \;\;\; &   
	\mathcal{F}(\boldsymbol{\theta}) =  \frac{1}{\textsf{D}} \sum_{i = 1}^{\textsf{D}}\mathcal{J}(f( \mathcal{X}_i|\boldsymbol{\theta}),\mathcal{Y}_i  )  ,
	%	 \nonumber\\
	%	\subjectto& f( \mathcal{X}_i|\boldsymbol{\theta}) = \mathcal{Y}_i,\;\;\; i=1,\dots, \textsf{D},
	\end{align}
	where the non-linear mapping between the input and output is constructed via  $f( \mathcal{X}_i|\boldsymbol{\theta})$ by minimizing the empirical loss $\mathcal{F}(\boldsymbol{\theta})$ over the learnable parameters $\boldsymbol{\theta}$ and the loss function $\mathcal{J}(\cdot)$ is the mean-squared-error (MSE) between the label data $\mathcal{Y}_i$ and the prediction $f( \mathcal{X}_i|\boldsymbol{\theta})$ of the learning model over the whole dataset, i.e., %defined by the learning model as
	\begin{align}
	\mathcal{J}(f(\mathcal{X}_i|\boldsymbol{\theta}),\mathcal{Y}_i) =  \| f( \mathcal{X}_i|\boldsymbol{\theta}) - \mathcal{Y}_i  \|_\mathcal{F}^2.
	\end{align}
	In CL, (\ref{lossML}) is solved at a PS, which collects the local datasets $\mathcal{D}_k, k \in \mathcal{K}$ from the clients. 
	
	On the other hand, transmission of datasets is avoided in FL %which are kept at the clients, instead only the model updates (gradients) are sent to the PS. Consequently, 
	by considering the following optimization problem 
	%	\textcolor{red}{is this problem correct? if the individual members of the sum in the objective are subjected to the equality constraint, then what is there to minimize?}
	\begin{align}
	\label{lossFL}
	\minimize_{\boldsymbol{\theta}} \;\;\; &   
	\frac{1}{K} \sum_{k=1}^{K}\mathcal{F}_k(\boldsymbol{\theta}), \nonumber\\
	\subjectto& \mathcal{F}_k(\boldsymbol{\theta}) = \frac{1}{\textsf{D}_k} \sum_{i = 1}^{\textsf{D}_k}\mathcal{J}(f( \mathcal{X}_i^{(k)}|\boldsymbol{\theta}),\mathcal{Y}_i^{(k)}  )  , 
	%	\nonumber\\
	%	&f( \mathcal{X}_i^{(k)}|\boldsymbol{\theta}) = \mathcal{Y}_i^{(k)},\; i=1,\dots, \textsf{D}_k, k \in \mathcal{K},
	\end{align}
	for which the learning model is trained over the local datasets $\mathcal{D}_k$ with input-output pair $(\mathcal{X}_i^{(k)},\mathcal{Y}_i^{(k)})$ and $\textsf{D}_k = |\mathcal{D}_k|$.

	\section{Hybrid Federated and Centralized Learning}
	\label{sec:HCFL}
	In this section, we introduce the proposed HFCL framework by taking into account the computational capability of the clients so that all of the clients can contribute to the learning task with their datasets regardless of their ability to compute model parameters. In ML tasks, training a model requires huge computational power to compute the model parameters. This requirement cannot always be satisfied by the computational capability of the client devices. In order to train the ML model effectively taking into account the computational capability of the clients, a hybrid training framework is introduced in this work. We assume that only a portion of the clients with sufficient computational power performs FL, while the remaining clients, which suffer from computational capability, send their datasets to the PS for model computation, as illustrated in Fig.~\ref{fig_Diag}.
	
	In order to train the learning model, the minimization of $\mathcal{F}(\boldsymbol{\theta})$ is carried out iteratively through gradient descent (GD). Denote the model parameters at the $t$-th communication round/iteration as $\boldsymbol{\theta}_t$, $t = 1,\dots,T$, where $T$ denotes the number of iterations to reach the convergence. Then, the $(t+1)$-th iteration of GD yields
	\begin{align}
	\label{eq:update1}
	\boldsymbol{\theta}_{t+1} = \boldsymbol{\theta}_t - \eta_t \mathbf{g}(\boldsymbol{\theta}_t),
	\end{align}
	where $\eta_t$ is the learning rate and
	\begin{align}
	\mathbf{g}(\boldsymbol{\theta}_t)= \nabla_{\boldsymbol{\theta}} \mathcal{F}(\boldsymbol{\theta}_t) =\frac{1}{\textsf{D}} \sum_{i=1}^{\textsf{D}}\nabla_{\boldsymbol{\theta}} \mathcal{J} (f ( \mathcal{X}_i|\boldsymbol{\theta}_t), \mathcal{Y}_i),
	\end{align}
	denotes the \emph{full} or \emph{batch} gradient vector in $\mathbb{R}^P$, where $P$ is the total number of learnable model parameters.
	
	For large datasets, GD is computationally inefficient. A better alternative is stochastic GD (SGD), where $\mathcal{D}$ is partitioned into $M_B$ mini-batches as $\mathcal{D} = \bigcup_{m\in \mathcal{M}_B} \mathcal{D}_m$, for $\mathcal{M}_B = \{1,\dots, M_B\}$~\cite{FL_QSGD}. Then,   $\boldsymbol{\theta}_t$ is updated by
	\begin{align}
	\label{gradientCL}
	\boldsymbol{\theta}_{t+1} = \boldsymbol{\theta}_t - \eta_t \mathbf{g}_{\mathcal{M}_B}(\boldsymbol{\theta}_t),
	\end{align}
	where ${\mathbf{g}}_{\mathcal{M}_B}(\boldsymbol{\theta}_t) =\frac{1}{M_B} \sum_{m=1}^{M_B} \mathbf{g}_m(\boldsymbol{\theta}_t) $ includes the contribution of gradients computed over $\{\mathcal{D}_m\}_{m\in \mathcal{M}_B}$ as 
	\begin{align}
	\mathbf{g}_m(\boldsymbol{\theta}_t) = \frac{1}{\textsf{D}_m} \sum_{i=1}^{\textsf{D}_m}\nabla_{\boldsymbol{\theta}} \mathcal{J} (f ( \mathcal{X}_{m,i}|\boldsymbol{\theta}_t), \mathcal{Y}_{m,i}) ,
	\end{align}
	where $(\mathcal{X}_{m,i},\mathcal{Y}_{m,i})$ denotes the $i$-th input-output pair for the $m$-th mini-batch and ${\textsf{D}_m} = |\mathcal{D}_m|$ is the mini-batch size. 
	
	The gradient term ${\mathbf{g}}_{\mathcal{M}_B}(\boldsymbol{\theta}_t)$ satisfies $\mathbb{E}\{ {\mathbf{g}}_{\mathcal{M}_B}(\boldsymbol{\theta}_t) \} = \nabla_{\boldsymbol{\theta}} \mathcal{F}(\boldsymbol{\theta}_t)$ and, therefore, SGD provides the minimization of the empirical loss by partitioning the dataset into $M_B$ mini-batches and accelerates the learning process, which is known as mini-batch learning~\cite{FL_QSGD}. Employing SGD in parallel among several devices allows us to compute the gradients on devices and aggregate them in the PS, which is known as FL~\cite{FL_Gunduz}. 
	
	%However, FL-based training demands huge computational power to compute the gradient information, which cannot always be satisfied by the computational capability of the client devices. Thus, 
	In our HFCL approach, %only a portion of the clients perform FL while the remaining clients, which suffer from computational capability, transmit their datasets to the PS for gradient computation (Fig.~\ref{fig_Diag}). By exploiting the computational resources of the clients in the hybrid architecture, 
	denote the group of active and passive clients that employ gradient computation at the PS and device levels as the index sets $\mathcal{L} = \{1,\dots,L\}$ and $\bar{\mathcal{L}} = \mathcal{K}\setminus  \mathcal{L} = \{L+1,\dots,K\}$, respectively. Then, (\ref{eq:update1}) becomes 
	\begin{align}
	\label{eq:update2}
	\boldsymbol{\theta}_{t+1} = \boldsymbol{\theta}_t - \eta_t \hspace{-2pt} \bigg( \underbrace{\frac{1}{L}\sum_{k\in \mathcal{L}} \mathbf{g}_k(\boldsymbol{\theta}_t)}_{\mathrm{On \;Server}}   + \underbrace{\frac{1}{K-L}\sum_{k\in \bar{\mathcal{L}}} \bar{\mathbf{g}}_k(\boldsymbol{\theta}_t)}_{\mathrm{On\; Device}} \bigg) \hspace{-2pt},
	\end{align}
	%We further call the clients in $\mathcal{L}$ and $\bar{\mathcal{L}}$ are \emph{passive} and \emph{active} clients, respectively, where the gradient-transmitting clients are defined as active. 
	the computation of $\mathbf{g}_{k\in \mathcal{L}}(\boldsymbol{\theta}_t)$ is done via mini-batch learning as $\mathbf{g}_{k\in \mathcal{L}}(\boldsymbol{\theta}_t) = \frac{1}{M_B}\sum_{m=1}^{M_B} \mathbf{g}_{m,k\in \mathcal{L}}(\boldsymbol{\theta}_t)$ where  $\mathbf{g}_{m,k\in \mathcal{L}}(\boldsymbol{\theta}_t) = \nabla_{\boldsymbol{\theta}} \mathcal{J} (f ( \mathcal{X}_{m,i}^{(k)}|\boldsymbol{\theta}_t), \mathcal{Y}_{m,i}^{(k)})$ considering that the PS has access to the dataset $\mathcal{D}_{k\in \mathcal{L}}$. The gradients corresponding to the active clients $\bar{\mathbf{g}}_{k\in \bar{\mathcal{L}}}$ are collected at the PS through a noisy wireless channel as
	\begin{align}
	\bar{\mathbf{g}}_{k\in \bar{\mathcal{L}}}(\boldsymbol{\theta}_t) = Q_B(\mathbf{g}_{k\in \bar{\mathcal{L}}}(\boldsymbol{\theta}_t)) + \mathbf{w}_{{k\in \bar{\mathcal{L}}},t},
	\end{align}
	where $Q_B(\cdot)$ represents the quantization operator with $B$ bits and $\mathbf{w}_{{k\in \bar{\mathcal{L}}},t}\in \mathbb{R}^P$ denotes the noise term added onto $Q_B(\mathbf{g}_{k\in \bar{\mathcal{L}}}(\boldsymbol{\theta}_t))$ at the $t$-th iteration. 
	
	Without loss of generality, we assume that $\mathbf{w}_{{k\in \bar{\mathcal{L}}},t}$ obeys normal distribution, i.e., $\mathbf{w}_{{k\in \bar{\mathcal{L}}},t}\sim \mathcal{N}(0,\sigma_{\boldsymbol{\theta}}^2\mathbf{I}_P)$ with variance $\sigma_{\boldsymbol{\theta}}^2$ and the signal-to-noise-ratio (SNR) in gradient transmission is given by  $\mathrm{SNR}_{\boldsymbol{\theta}} = 20\log_{10}\frac{||\mathbf{g}_{{k\in \bar{\mathcal{L}}}}(\boldsymbol{\theta}_t)||_2^2}{\sigma_{\boldsymbol{\theta}}^2  }  $.	The same amount of noise is also added onto $\mathcal{D}_{k\in \mathcal{L}}$ during dataset transmission for passive clients. Once the model aggregation is completed during model training, the PS returns the updated model parameters $\boldsymbol{\theta}_{t+1}$ to only the active clients. 
	
	%	To reduce the oscillations due to the gradient averaging, parameter update is performed by using a momentum parameter $\gamma$ which allows us to ``moving-average'' the gradients~\cite{FL_QSGD,elbir2020FL_HB}. Finally, the parameter update with momentum is given by
	%	\begin{align}
	%	\label{gradientUpdateWithmomentum}
	%	\boldsymbol{\theta}_{t+1} = \boldsymbol{\theta}_t - \eta_t  \frac{1}{K} \sum_{k=1}^{K} \mathbf{g}_k(\boldsymbol{\theta}_t) + \gamma (\boldsymbol{\theta}_t - \boldsymbol{\theta}_{t-1}).
	%	\end{align}
	
	\subsection{Communication Delay During Model Training}
	The bandwidth resources need to be optimized to reduce the  latency of the transmission of  both $\mathbf{g}_k(\boldsymbol{\theta})$ ($k\in \bar{\mathcal{L}}$) and $\mathcal{D}_k$ ($k\in \mathcal{L}$) to the PS during training. Let $\tau_k$ be the communication time for the $k$th client to transmit its either dataset ($\mathcal{D}_k$ for $k\in \mathcal{L}$) or model updates ($\mathbf{g}_k(\boldsymbol{\theta})$ for $k\in \bar{\mathcal{L}}$), and can defined as $	\tau_k = \frac{d_k}{R_k},$
	%	\begin{align}
	%	\label{latency}
	%	\tau_k = \frac{d_k}{R_k},
	%	\end{align}
	where $d_k$ denotes the number of dataset symbols to transmit and $R_k = B_k \ln (1 + \textsf{SNR}_k )$ is the achievable transmission rate, for which $B_k$ and $\textsf{SNR}_k$, respectively, denote the allocated bandwidth and the signal-to-noise ratio (SNR) for the $k$th client. 	The PS solves $\min_{B_k} \max_{k\in \mathcal{K}} \hspace{10pt} \tau_k $ to optimize the bandwidth allocation by minimizing the maximum communication delay. This is because the model aggregation in the PS can be performed only after the completion of the slowest transmission for $k\in \mathcal{K}$.  Although $R_k$ can vary for $k\in \mathcal{K}$, $d_k$ differentiates more significantly than $R_k$ between the passive (i.e., $k\in \mathcal{L}$) and active (i.e., $k\in \bar{\mathcal{L}}$) clients~\cite{latencyMinFL}. Depending on the client type, $d_k$ can be given by
	\begin{align}
	\label{latency2}
	d_k = \left\{\begin{array}{ll}
	P, & k \in \bar{\mathcal{L}} \\
	\textsf{d}_k, & k\in \mathcal{L}
	\end{array}\right.,
	\end{align}
	which is fixed to the number of model parameters $P$ for the active clients and to $\textsf{d}_k = D_k(U_xV_x + U_yV_y)$ for $D_k$ input ($\in \mathbb{R}^{U_x \times V_x}$) and output ($\in \mathbb{R}^{U_y \times V_y}$) dataset samples. Since the dataset size is usually larger than the number of model parameters in ML applications, i.e., $\textsf{d}_{k\in \mathcal{L}} > P$~\cite{elbir2020FL_HB,FL_Gunduz,fl_spm_federatedLearning}, the dataset transmission of the passive clients is expected to take longer than the model transmission of the active clients, i.e., $\tau_{k\in \mathcal{L}} > \tau_{k\in \bar{\mathcal{L}}}$~\cite{latencyMinFL}. Previous FL-based works reported that $\tau_{k\in \mathcal{L}}$ can be approximately $10$ times  longer than $\tau_{k\in \bar{\mathcal{L}}}$~\cite{elbir2020_FL_CE,elbir2020FL_HB}. This introduces a significant delay especially at the beginning of the training. Because, the HFCL problem in (\ref{eq:update2})  can be performed only if $\mathcal{D}_{k\in \mathcal{L}}$ is collected at the PS for the first iteration. 
	To circumvent this issue and keep the training continue, we propose the SDT approach in the following.
	
	\subsection{HFCL With Sequential Data Transmission}
	The HFCL approach described above performs model aggregation only after the dataset transmission for $k\in \mathcal{L}$ is completed when $t =1$. This causes delays during model training. In order to circumvent this problem, we propose HFCL-SDT that partitions the local dataset of the passive clients into small blocks and transmits them to the PS sequentially without any delay. A conventional way is to partition the dataset $\textsf{D}_{k\in \mathcal{L}}$ into $ N = \frac{\textsf{D}_k}{P}$ blocks, where the size of the transmitted symbols for each communication round is equal to $P$ for both passive and active  clients. For non-integer $N$, $N=\lceil \frac{\textsf{D}_k}{P} \rceil$. 
	
	Then, compute $\mathbf{g}_{k\in \mathcal{L}}(\boldsymbol{\theta}_t) = \frac{1}{M_B}\sum_{m=1}^{M_B} \mathbf{g}_{m,k\in \mathcal{L}}(\boldsymbol{\theta}_t)$, where
	\begin{align}
	\label{eq:update3}
	\hspace{-7pt} \mathbf{g}_{m,k\in \mathcal{L}}(\boldsymbol{\theta}_t) \hspace{-3pt}=\hspace{-3pt} \left\{\begin{array}{ll} \hspace{-7pt}    \frac{1}{tP} \sum_{i=1}^{tP } \hspace{-2pt} \nabla_{\boldsymbol{\theta}} \mathcal{J} (f ( \mathcal{X}_{m,i}^{(k)}|\boldsymbol{\theta}_t), \mathcal{Y}_{m,i}^{(k)}) , &\hspace{-9pt} t \leq N\\ \hspace{-7pt} 
	\frac{1}{\textsf{D}_k} \sum_{i=1}^{\textsf{D}_k }\hspace{-2pt} \nabla_{\boldsymbol{\theta}} \mathcal{J}(f ( \mathcal{X}_{m,i}^{(k)}|\boldsymbol{\theta}_t), \mathcal{Y}_{m,i}^{(k)})  , &\hspace{-9pt} t > N\end{array}\right.\hspace{-8pt}.
	\end{align}
	Note that the size of the training dataset for $\mathbf{g}_{k\in \mathcal{L}}(\boldsymbol{\theta}_t)$ grows as $t \rightarrow N$. It is, however, constant for $t > N$ when the size of the transmitted dataset is fixed at $P$ because the collected blocks of the dataset are stored at the PS for $t\leq N$. Consequently, HFCL-SDT mitigates transmission delay but exhibits the same communication overhead as HFCL. Furthermore, HFCL-SDT performs better than HFCL because smaller datasets imply quick learning of the features in the data at the beginning of the training.
	
	%\section{Communication Overhead in HFCL}
	% NO need for a such a short section
	Communication overhead is measured by the number of transmitted symbols during model training~\cite{elbir2020FL_HB,FL_Gunduz,elbir2020cognitive,fl_spm_federatedLearning}. For CL ($\mathcal{T}_{\mathrm{CL}}$), this overhead is the number of symbols used to transmit datasets. The same for FL ($\mathcal{T}_{\mathrm{FL}}$) is proportional to the number of communication rounds $T$ and  model parameters $P$. Denote $\bar{\textsf{D}} = \sum_{k}\bar{\textsf{D}}_k $ as the number of symbols of the whole dataset. The communication overheads of CL, FL and HFCL are, respectively, 
	\begin{align}
	\mathcal{T}_{\mathrm{CL}} &= \bar{\textsf{D}}, \\
	\mathcal{T}_{\mathrm{FL}} &= 2TPK, \\
	\mathcal{T}_{\mathrm{HFCL}} &= L\bar{\textsf{D}}_{k\in \mathcal{L}} + 2TP(K-L),
	\end{align}
	where $\mathcal{T}_{\mathrm{HFCL}}$ includes the transmission of dataset of $L$ passive clients and gradients of $K-L$ active clients. In general, FL has lower communication overhead than CL~\cite{elbir2020FL_HB,FL_Gunduz,elbir2020cognitive,elbir2020_FL_CE,fl_spm_federatedLearning}. Hence, it follows that $\mathcal{T}_{\mathrm{FL}}\leq \mathcal{T}_{\mathrm{HFCL}}\leq \mathcal{T}_{\mathrm{CL}}$.

	%%-----------------------------------------------------
	\begin{figure}[t]
		\centering
		{\includegraphics[draft=false,width=.95\columnwidth]{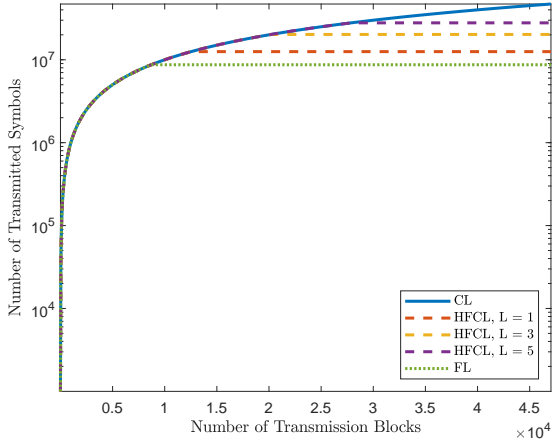} } 
		\caption{Communication overhead comparison.
		}
		%			\vspace*{-5mm}
		\label{fig_TO}
	\end{figure}
	%%-----------------------------------------------------

	\section{Numerical Simulations}
	\label{sec:sim}
	We evaluated the performance of our HFCL approach using the MNIST dataset~\cite{mnistlecun2010mnist} comprising $28\times 28$ gray-scale images of handwritten digits with $10$ classes. The number of symbols on the whole dataset is $\bar{\textsf{D}} = 28^2\cdot60,000 \approx 47\times 10^6$. During model training, the dataset is partitioned into $K=10$ blocks, each of which is available at the clients that are independently and identically distributed. Further, we train a CNN with two convolutional layers with $5\times 5$@$128$ and $3\times 3$@$128$ spatial filters. Thus, we have $P=128(5^2 + 3^2) = 4,352$. The validation data of MNIST dataset includes $10,000$ images and it is used for performance comparison for the competing algorithms. The learning rate is selected as $0.001$ and the mini-batch size is $128$ for CL. The loss function was the cross-entropy cost as $ -\frac{1}{{\textsf{D}}} \sum_{i = 1}^{{\textsf{D}}} \sum_{c = 1}^{\bar{C}}  \bigg[ \mathcal{Y}_i^{(c)} \ln \hat{\mathcal{Y}}_i^{(c)} + (1 - \mathcal{Y}_i^{(c)}) \ln (1 - \hat{\mathcal{Y}}_i^{(c)})  \bigg],$	where $\{\mathcal{Y}_i^{(c)}, \hat{\mathcal{Y}}_i^{(c)}  \}_{i= 1, c=1}^{{\textsf{D}}, \bar{C}}$ is the true and predicted response for the classification layer  with $\bar{C}=10$ and the classification accuracy is $	\mathrm{Accuracy}(\%) = \frac{\textsf{U}}{\textsf{D}}\times 100,$ in which the model identified the image class correctly $\textsf{U}$ times.

	%%-----------------------------------------------------
	\begin{figure}[t]
		\centering
		{\includegraphics[draft=false,width=\columnwidth]{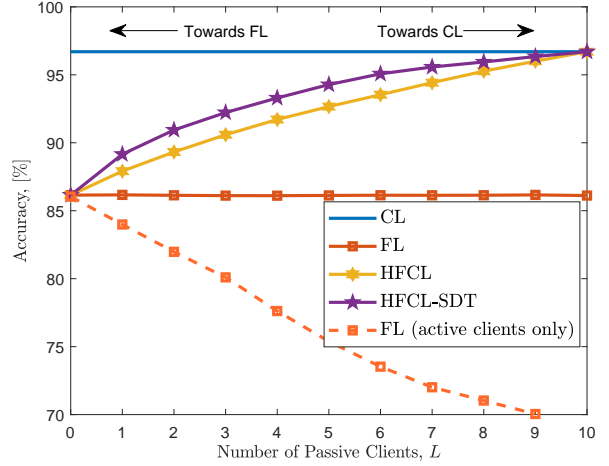} } 
		\caption{Classification accuracy versus  $L$ when $\mathrm{SNR}_{\boldsymbol{\theta}}=20$ dB and  $B= 5$.
		}
		%			\vspace*{-5mm}
		\label{fig_acc_L}
	\end{figure}
	%%-----------------------------------------------------
	Fig.~\ref{fig_TO} depicts the communication overhead of CL, FL and HFCL for  $L=\{0, 1,3,5,7,10\}$. During model training, $1000$ data symbols are transmitted at each transmission block. It takes approximately $47\times 10^3$ transmission blocks to complete CL-based training while FL demands approximately  $8.5\times 10^3$  data blocks (approximately $6$ times lower than that of CL). The communication overhead of HFCL (as well as HFCL-SDT) lies between CL and FL because it depends on the number of passive clients $L$ and approaches to $\mathcal{T}_\mathrm{CL}$ as $L\rightarrow K$.

	Fig.~\ref{fig_acc_L} shows the classification accuracy with respect to number of passive clients, $L$ when $B=5$ quantization bits are used and $\mathrm{SNR}_{\boldsymbol{\theta}}= 20$ dB. The proposed HFCL and HFCL-SDT approaches perform better than FL for $0< L < K$ because the collected gradients from the active clients are corrupted by wireless channel and quantization whose effects reduce as $L \rightarrow K$.  When $L=0$, HFCL and HFCL-SDT are identical to FL (all clients are active) whereas they perform identically as CL if $L=K=10$ (all clients are passive). The	HFCL-SDT provides higher accuracy than HFCL for $0< L < K$ because the former performs gradient computation on smaller datasets at the beginning of training for $t < N$ thus reaching higher accuracy levels quicker than the latter. %, which computes the gradient information on whole local dataset of the passive clients for $t < N$ leading to a slower convergence rate.
	%%-----------------------------------------------------
	\begin{figure}[t]
		\centering
		{\includegraphics[draft=false,width=.95\columnwidth]{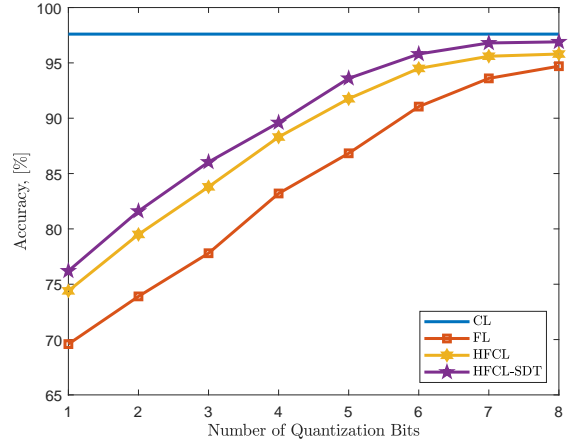} } 
		\caption{Classification accuracy versus $B$ when  $\mathrm{SNR}_{\boldsymbol{\theta}} = 20$ dB.
		}
		%			\vspace*{-5mm}
		\label{fig_acc_Q}
	\end{figure}
	%%-----------------------------------------------------

	%%-----------------------------------------------------
	\begin{figure}[t]
		\centering
		{\includegraphics[draft=false,width=.95\columnwidth]{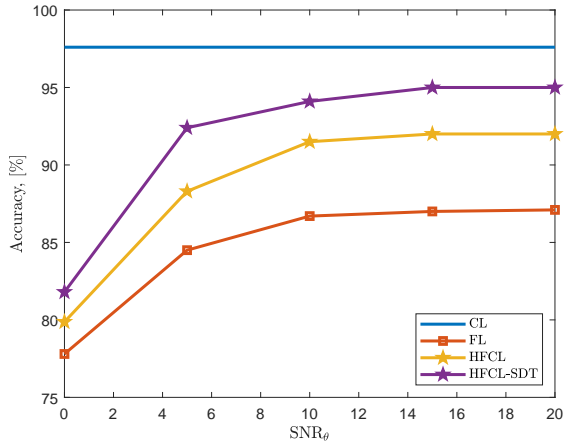} } 
		\caption{Classification accuracy versus $\mathrm{SNR}_{\boldsymbol{\theta}}$ when $Q = 5$.
		}
		%			\vspace*{-5mm}
		\label{fig_acc_SNRt}
	\end{figure}
	%%-----------------------------------------------------

	In Fig.~\ref{fig_acc_L}, we also present the performance of FL with active clients only, that is to say, the learning model is trained only on  the dataset of active clients, whereas it is tested on the whole dataset. We observe that FL becomes unable to learn the data as $L$ increases since the training is conducted only on the datasets of active clients. This shows the effectiveness of our HFCL approach, in which all of the clients participate the learning stage. It is worthwhile to note that when $L=K=10$, there will be no active clients as FL with only active clients cannot work. The performance loss as $L\rightarrow K$ is due to the absence of passive clients' datasets can be severe if the dataset is non-identically distributed because the active clients cannot learn the whole features in the dataset of other devices.

	The classification accuracy (Fig.~\ref{fig_acc_Q}) in terms of the quantization levels for $B \in [1,8]$ when $\mathrm{SNR}_{\boldsymbol{\theta}}= 20$ dB shows that both HFCL approaches perform better than FL and approach to CL as an increase in $B$ improves the resolution of the quantization operation. The classification performance in Fig.~\ref{fig_acc_SNRt} with respect to the noise level on the model parameters, i.e., $\boldsymbol{\theta}$ and $\mathbf{g}_k(\boldsymbol{\theta})$ compares the competing algorithms for $\mathrm{SNR}_{\boldsymbol{\theta}}\in [0, 20]$ dB. Here, at least $10$ dB noise level is required for reliable model training for all approaches. While the same level of noise is also added onto the dataset of passive clients, its influence is more apparent on the gradients since it directly affects the learning performance.

	\section{Summary}
	We introduced a hybrid federated and centralized learning (HFCL) approach for distributed machine learning tasks. The proposed approach is helpful if a part of the edge devices lack computational capability for gradient computation during model training. In order to train the learning model collaboratively, only active devices with sufficient computational capability perform gradient computation on their local datasets whereas the remaining passive devices transmit their local datasets to the PS. %, in which the gradient computation is performed. 
	The delays arising from the transmission of local datasets during training of the size of the passive clients are large are mitigated by HFCL-SDT. %, wherein better classification accuracy can be achieved due to SDT. As future work, we reserve to study the latency considerations of the HFCL-based training techniques and the application of such approaches for fundamental wireless communications problems, such as channel estimation~\cite{elbir2020_FL_CE}, resource allocation and beamforming~\cite{elbir2020FL_HB}.
	
	\section*{Acknowledgements}
	S. Coleri acknowledges the support of the Scientific and Technological Research Council of Turkey (TUBITAK) EU CHIST-ERA grant 119E350. A. M. Elbir acknowledges the support of TUBITAK. 
	
	\balance
	\bibliographystyle{IEEEtran}
	\bibliography{IEEEabrv,references_085}

	%TC:ignore
	%	\begin{IEEEbiographynophoto} {Ahmet M. Elbir} (IEEE Senior Member) received the B.S. degree with Honors from Firat University in 2009 and the Ph.D. degree from Middle East Technical University (METU) in 2016, both in electrical engineering. He is the recipient of 2016 METU best Ph.D. thesis award for his doctoral studies. He serves as an Associate Editor for IEEE Access since 2018. Currently, he is a visiting postdoctoral researcher at Koc University, Istanbul, Turkey; and senior researcher at Duzce University, Duzce, Turkey. His research interests include array signal processing, sparsity-driven convex optimization, signal processing for communications and deep learning for array signal processing.
	%	\end{IEEEbiographynophoto}
	%	
	%TC:endignore
	
\end{document}